\documentclass[letterpaper,journal]{IEEEtran}
\usepackage{amsmath,amssymb,bm}
\usepackage{graphicx}
\usepackage{booktabs}
\usepackage{multirow} 
\usepackage[table]{xcolor}
\usepackage{textcomp}
\usepackage{url}
\usepackage{cite}
\usepackage[font=footnotesize,labelfont=bf]{caption}
\usepackage{subcaption}
\usepackage{cuted} 

\definecolor{viyellow}{RGB}{255,221,51}
\definecolor{navorange}{RGB}{255,140,20}
\definecolor{imagenta}{RGB}{210,120,220}
\definecolor{ourblue}{RGB}{60,150,230}

\newcommand{\hl}[2]{{\setlength{\fboxsep}{0pt}\colorbox{#1}{\strut #2}}}
\begin{document}

\title{Learning to Navigate with Minimal Parameters: Decomposing Visual Navigation Through Closed-Form Geometric Interfaces}

\author{
    Edward Beng Wai Tan$^{*,1}$, Siew-Kei Lam$^{1}$%
    \thanks{$^{1}$College of Computing and Data Science, Nanyang Technological University, Singapore.}%
    \thanks{$^{*}$Corresponding author.}%
}

\maketitle
\thispagestyle{empty}
\pagestyle{empty}

\begin{strip}

  \centering
  \begin{minipage}[b]{0.49\textwidth}
    \centering
    {\includegraphics[width=\textwidth]{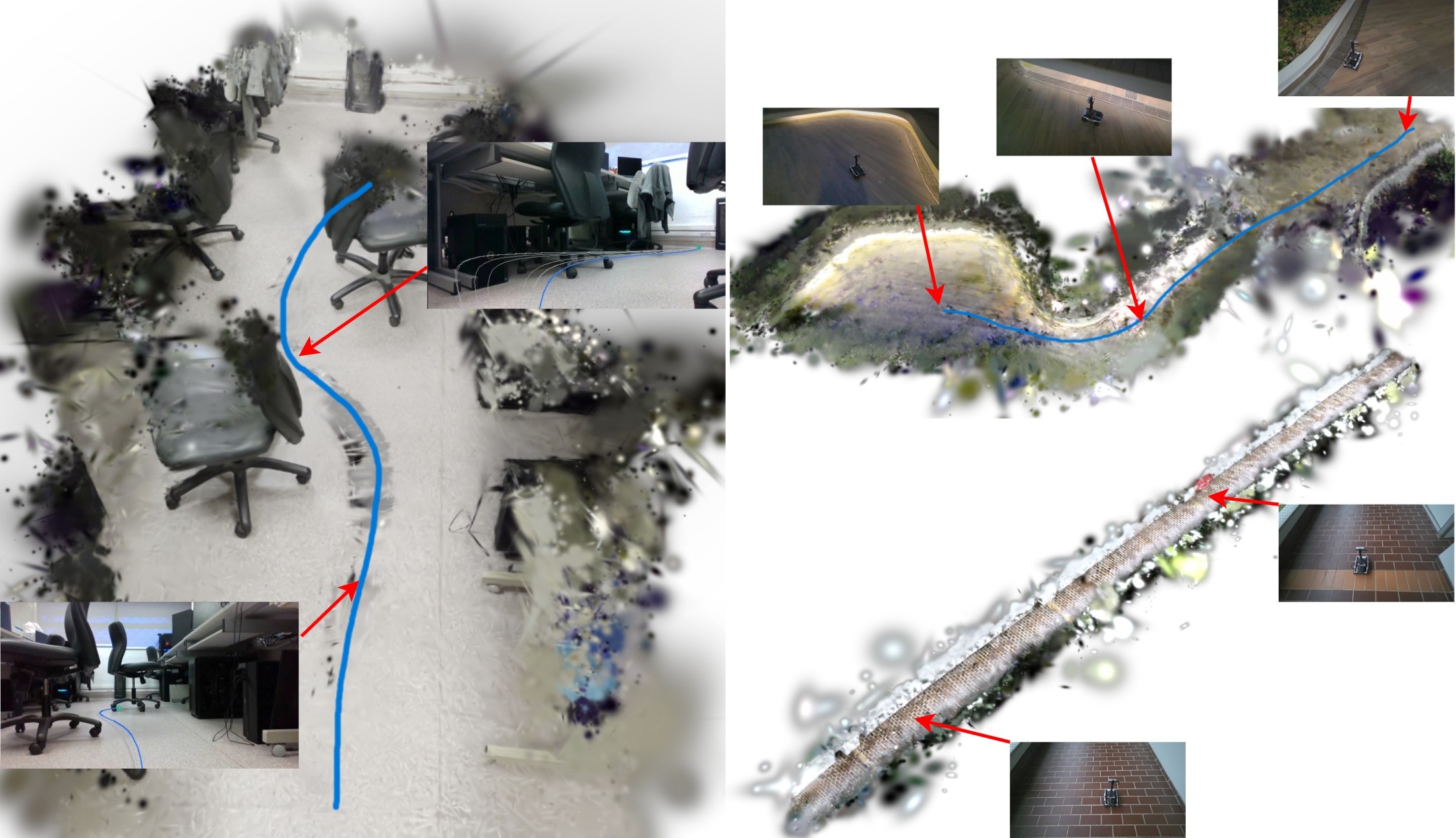}}
  \end{minipage}\hfill
  \begin{minipage}[b]{0.49\textwidth}
    \centering

    {\includegraphics[width=\textwidth]{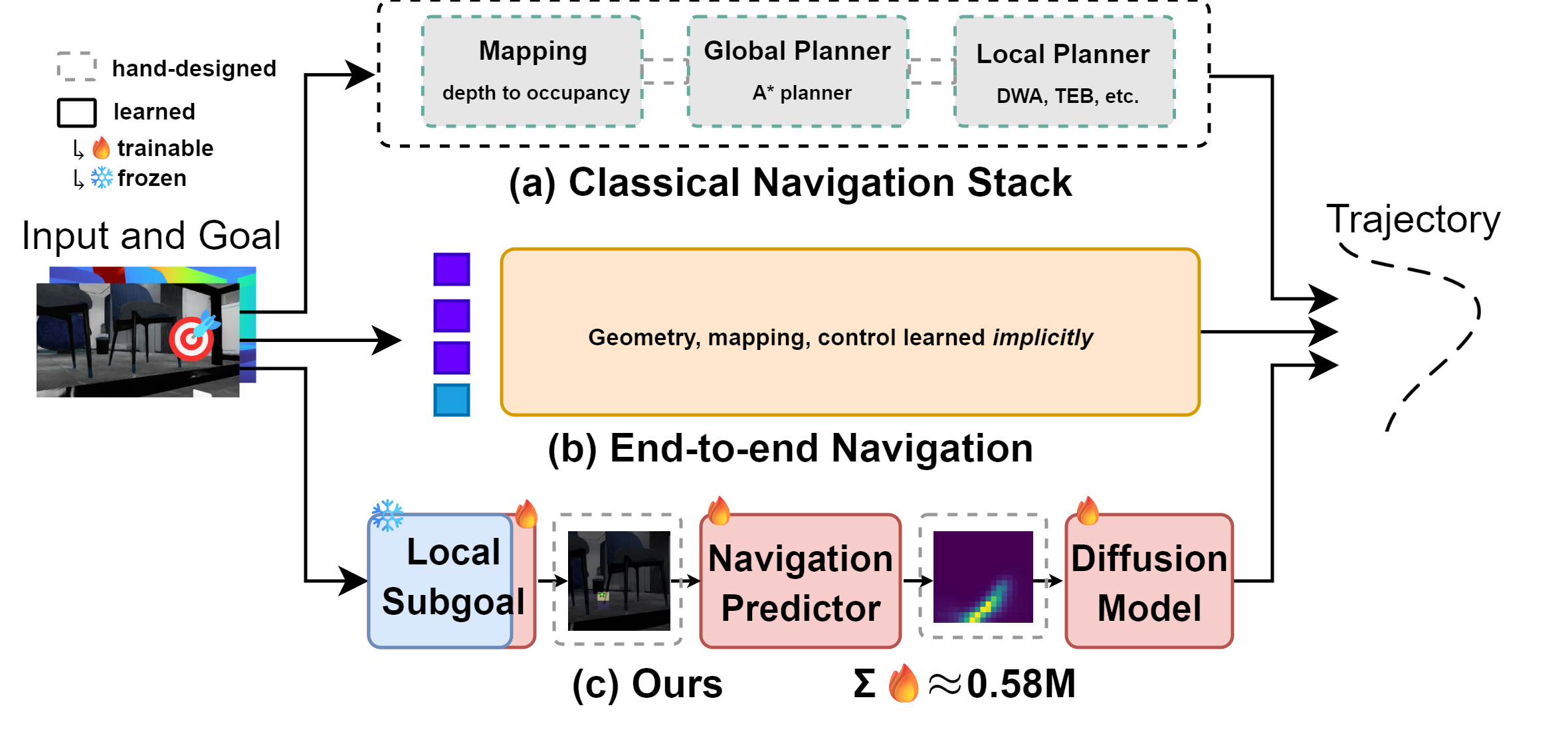}}
  \end{minipage}

  \captionof{figure}{\textbf{Real-world demonstrations (left).} We test our system in both indoor and outdoor environments under varying illumination conditions. The indoor trials are conducted in a typical office environment, while the outdoor trials are in corridors and public spaces.  
  \textbf{Comparison of navigation methods (right). }(a) Classical navigation algorithms are efficient but hand-designed heuristics can limit performance, (b) end-to-end navigation models learn geometry, mapping, and control implicitly, at the cost of large parameter and scaling cost, (c) our method retains closed-form solutions for operations with known structure, allowing the network to learn the rest of the navigation task with only 0.58M trainable parameters.
  }
  \label{fig:hero}
\end{strip}

\begin{abstract}
    Visual navigation policies have grown to hundreds of millions of parameters trained on billions of frames, with geometry, mapping, and control learned implicitly. We propose a decomposed point-goal navigation system in which operations with known closed-form structure, such as projective geometry, occupancy, and coordinate transforms, are computed analytically and serve as interfaces between three small learned modules: an egress predictor that grounds the episode goal as a local subgoal in the current view, a navigation predictor that estimates a goal-conditioned posterior over where trajectories travel, and an endpoint-pinned residual diffusion generator that samples trajectory shapes from this posterior. Only 0.58M out of 23M parameters are trained, on 44k frames, in under one GPU-hour. Across 6060 point-goal episodes in 60 environments, the system attains competitive success rates with the lowest collision rate among evaluated methods. We further show that under this decomposition, the frozen image encoder can be replaced by a 0.54M MobileNetV2 at a -2.0 SR cost, bringing the full system under 1.2M parameters. It also transfers to no-goal exploration by retraining only the 123k-parameter egress head, and its failure modes under sensor corruption are transparent and analytically correctable. We deploy and evaluate the system zero-shot on a low-cost UGV, running navigation and localization on a Jetson Orin Nano in real-time.
\end{abstract}


\section{Introduction}

Visual navigation is a challenging task, requiring robots to navigate to objectives within previously unseen environments. The prevailing route to these capabilities has largely been through
scale: reinforcement learning policies trained on billions of steps
\cite{Wijmans2020DD-PPO:}, imitation policies trained on large cross-embodiment
corpora \cite{shah2022gnm, shah2023vint}, and more recently, diffusion-based
foundation policies trained for trajectory generation on large-scale simulated data \cite{caiNavDPLearningSimtoReal2025}. The capability of these models has improved steadily, but at the cost of parameter count, dataset size, and inference cost.  These data-driven advancements often bundle many emergent abilities, such as multiple goal types, embodiment awareness, and implicit spatial mapping.

However, many resource-constrained robots, particularly those which require on-device inference, cannot afford to run these large policies at sufficiently high speed, yet would benefit from the task-specific navigation performance of these models. Motivated by this, we ask the question: \textit{for a single task family such as point-goal navigation, how much needs to be learned when known geometric structure is explicitly used as an interface between learned operators?}

To address this, we propose a hybrid approach as shown in Figure~\ref{fig:hero} (right): combining the strengths of classical methods in performing closed-form operations (e.g., projective geometry, occupancy) as the interfaces to reduce the learning load on the neural components, while learning the operations between them. To that end, we factor the point-goal visual navigation task into three distinct operations, each represented by a small, trained operator. The \emph{egress predictor} learns a local point-goal given the final point-goal, the \emph{navigation predictor} learns the goal-conditioned map given the depth-derived Bird's-Eye-View (BEV) map, and the \emph{generator} learns the shape of the trajectory from the outputs of the other two predictors.

Each sub-task as described above is relatively simple, and is thus learned using a total of only 0.58M \textit{trainable} parameters (with 22.7M in total, including the \emph{fully} frozen image encoder), trains on 44k frames in under one GPU-hour, while performing competitively to significantly larger models on the InternRobotics point-goal benchmarks, trailing the
135.7M-parameter NavDP by a few points while using significantly fewer trainable parameters. We also show that the image encoder can be made significantly smaller with minimal performance reduction, for a \textit{total} system parameter count of 1.2M. Beyond the low parameter count, we show that our proposed decomposition generalizes well to other tasks (e.g., no-goal navigation), is relatively robust under sensor corruption, and navigates with few collisions.

We state our contributions below:
\begin{itemize}
    \item We frame point-goal navigation as being realizable as a composition of simpler sub-tasks, and by learning only minimal parameters, we demonstrate competitive performance without significant data or parameter scaling.
    \item We realize the sub-tasks as an egress predictor, a navigation
    field posterior, and an endpoint-pinned residual diffusion generator,
    connected by closed-form geometry.
    \item We show through collision analysis, sensor-corruption studies, and
    interface ablations that the proposed decomposition is robust to real-world conditions, and is useful beyond the particular point-goal task family.
\end{itemize}

\section{Related Work}

\noindent\textbf{Navigation at scale.}
The dominant route to recent point-goal navigation has been to scale end-to-end policies: DD-PPO reaches near-perfect simulated success from billions of frames \cite{Wijmans2020DD-PPO:}, GNM \cite{shah2022gnm} and ViNT \cite{shah2023vint} generalize across embodiments through large cross-robot corpora, and NavDP \cite{caiNavDPLearningSimtoReal2025} gains cross-embodiment awareness through large-scale simulated data. LoGoPlanner \cite{pengLoGoPlannerLocalizationGrounded2025} goes further, integrating localization and metric-awareness into a single navigation framework. In these 
works, geometric structure, projective relations and other operations are implicitly learned through data; part of the scale is spent on learning these tasks.

\noindent\textbf{Diffusion trajectory generation.}
Diffusion Policy established denoising as a representation for multimodal action distributions \cite{chi2023diffusionpolicy}, and navigation instances \cite{sridhar2023nomad, caiNavDPLearningSimtoReal2025} inherit both its expressiveness and its computational/parameter cost. Efficiency work within this line accelerates the sampler using neural prior initialization \cite{ren2025prior,luo2026stepnavstructuredtrajectorypriors}, and incorporating few-step bridge diffusion techniques \cite{luanRectifiedSchrodingerBridge2026} to accelerate generation. Other works tackle the data-efficiency problem, using output spline parameterization as inductive biases \cite{wangSanDPlannerSampleEfficientDiffusion2026}. These works largely tackle the data efficiency and generation efficiency problem, but these diffusion models still rely on learned RGB features as conditioning.

\noindent\textbf{Decomposed planners and subgoals.}
Decomposing navigation is not fundamentally new. Many classical planners are built on a global-to-local split, in which a global planner supplies waypoints to a reactive local method \cite{foxDynamicWindowApproach1997}, or learned hierarchical Reinforcement Learning (RL) \cite{lee2023adaptive} techniques which learn a set of low-level policies selectively deployed by a high-level policy, modular map-based agents whose learned global policy selects goals executed by an analytic planner \cite{NEURIPS2020_2c75cf26}, and more recently imperative planners in which a network predicts sparse waypoints refined by a differentiable optimizer over a geometric or semantic costmap \cite{roth2024viplanner, Yang-RSS-23}. Most recently, DualVLN \cite{weiGroundSlowMove2025} decomposes a different task, vision-language navigation, through intermediate pixel goals between a reasoning module and a trajectory generator, showing that explicit subgoal interfaces retain benefits even within data scaling methods. What these works have in common is being coupled through learning: imperative front-ends receive gradients through the optimizer \cite{Yang-RSS-23}, and DualVLN's subgoal is accompanied by a latent channel optimized jointly with the policy \cite{weiGroundSlowMove2025}. However, for recent learned methods, few works test the decomposition itself, but rather benefit from additional training objectives introduced at the interfaces. Motivated by this, we propose an architecture which addresses visual navigation using learnable modules, each of which perform a specific sub-task.

\begin{figure*}[t]
  \centering
  \includegraphics[width=\textwidth]{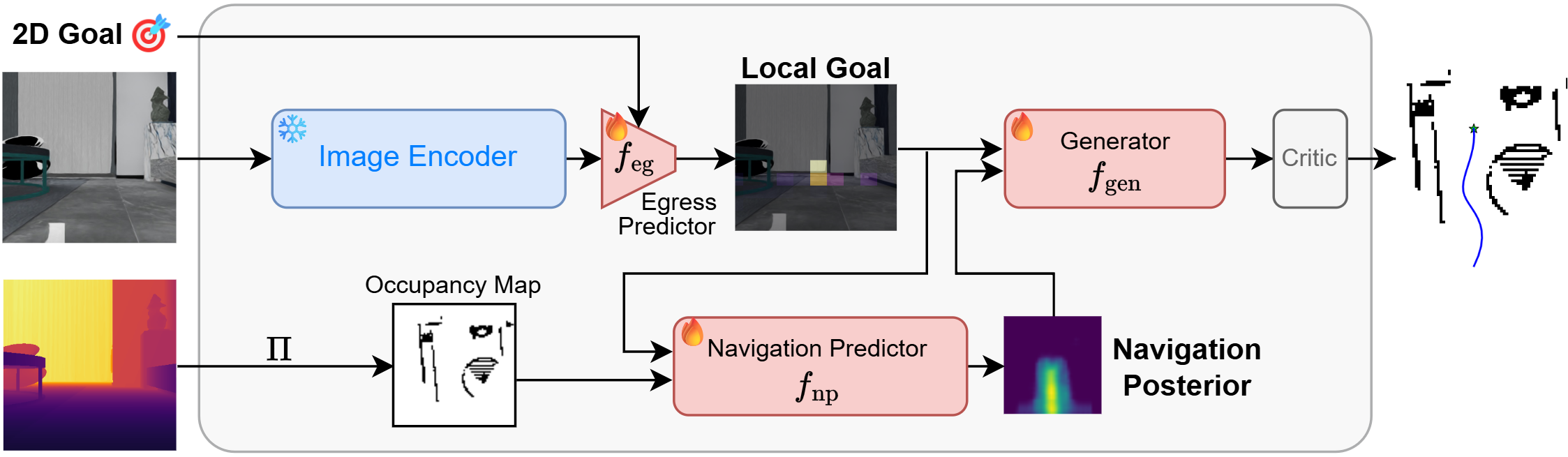}
  \caption{\textbf{System Architecture.} Our system consists of three learnable modules, each performing a distinct task. The \textit{egress predictor} $f_{\mathrm{eg}}$ learns a local navigation subgoal, the \textit{navigation predictor} $f_{\mathrm{np}}$ learns goal-conditioned spatial posterior, and the \textit{generator} $f_{\mathrm{gen}}$ predicts the trajectory shape between the current position and predicted subgoal. Modules operate in their native coordinate frames and exchange information through closed-form geometric transforms. }
  \label{fig:system}
\end{figure*}

\section{Method}

\subsection{Overall Architecture}
We define a navigation model composed of three distinct learnable components, as illustrated in Figure~\ref{fig:system}, each of which learns a specific task as described below:
\begin{itemize}
    \item The \textbf{egress predictor} takes as input the episode-level goal, and learns the pixel-goal target in the current view.
    \item The \textbf{navigation predictor} consumes the BEV occupancy map and the egress goal, and learns the field representing the posterior probability of where the trajectory should go.
    \item The \textbf{generator} predicts the shape of the trajectory whose endpoint is determined by the egress predictor.
\end{itemize}

Each learnable module explicitly operates within its own coordinate frame consistent with the corresponding subproblem: the egress predictor in the image frame, the navigation predictor and generator in the BEV-frame. The transformation is performed between modules in closed-form, and thus each module avoids expending learning capacity on trivially defined operations.

At each control cycle, our system takes as input the RGB image $I_t$ and the depth $D_t$. We first analytically calculate the occupancy map $O_t$ in BEV-frame from the unprojected depth, from which a costmap is derived as the combination of an exponential repulsion of the obstacle distance transform $d(x)$ with a bearing ramp $\rho(x)$ penalizing out-of-view mass:
\begin{equation}
C_t(x) \;=\; \max\!\big(e^{-d(x)/\lambda_d},\, \rho(x)\big).
\label{eq:cost}
\end{equation}

In subsequent operations, the egress predictor sees only the RGB image $I_t$, and the navigation predictor sees only the depth-derived occupancy map $O_t$. The generator receives neither input directly. The costmap $C_t$ is used as an auxiliary term for the generator, and also after generation to select the best candidate trajectory.

\subsection{Egress operator $f_{\mathrm{eg}}$: endpoint selection}
\label{subsec:egress}
We use a DepthAnythingv2 \cite{NEURIPS2024_26cfdcd8} ViT-S as the image encoder, which is \emph{fully} frozen, without any adapters, prompt tuning or other such fine-tuning methods. Thus, it learns no additional navigation-specific properties. In Section~\ref{subsec:generalize}, we also show that smaller encoders can be used with minimal accuracy reduction. The egress predictor is a small neural network conditioned on the image encoder features $\Phi$ and the goal $\hat G_t$ as:
\begin{equation}
z_t = f_{\mathrm{eg}}\big(\Phi(I_t),\, \hat G_t\big),
\end{equation}
where $z_t$ denotes a grid of logits. The predictor is trained to minimize the cross-entropy loss over a discrete patch grid:
\begin{equation}
\mathcal{L}_\mathrm{eg} = \mathbb{E}_{(\Phi(I_t), \hat G_t, c_t)\sim \mathcal{D}}[- \log p(c_t \mid \Phi(I_t), \hat G_t)],
\label{eq:unproj}
\end{equation}
where $c_t$ is the point on the rasterized image-frame where the path to $\hat G_t$ exits the current BEV window or passes behind objects, subject to a maximum distance truncation.
At inference time, the image-frame goal is equivalent to $\arg\max(z_t)$, and the target endpoint $g_t$ takes the 3D position of its ray-cast intersection with the ground plane.

\subsection{Navigation predictor $f_{\mathrm{np}}$}
The navigation predictor estimates the conditional distribution
\begin{equation}
P_t \;=\; P\big(\cdot \,\big|\, g_t, O_t\big),
\end{equation}
through a lightweight operator $f_{\mathrm{np}}(g_t, O_t)$ emitting logits in the BEV coordinate frame, and each discrete coordinate encodes a scalar value and direction.
The head emits $K$ recurrent posteriors $\{P_k\}_{k=1}^{K}$, one per refinement step.
The $k$-th map is supervised against its own target route, which anneals from
the straight-line prior to the ground-truth path,
\begin{equation}
\tau_k = (1-\alpha_k)\,\tau_{\mathrm{prior}} + \alpha_k\,\tau_{\mathrm{gt}},
\qquad \alpha_k = \tfrac{k}{K},
\end{equation}
from which we render a soft occupancy target $T_k=\mathrm{raster}(\tau_k)$ and a
per-cell tangent target $\theta^*_k$ (the path tangent at the nearest point of
$\tau_k$):
\begin{equation}
\ell_k(P_k,T_k) = \underbrace{H(T_k,P_k)}_{\text{route}}
\;+\; \lambda_\theta \underbrace{\big\langle T_k,1- \operatorname{cossim}(\hat\theta_k,\theta^*_k)\big\rangle}_{\text{heading}} ,
\end{equation}
where $H(\cdot, \cdot)$ is the cross-entropy and  $\langle A,B\rangle = \sum_x A(x)B(x)$ sums over BEV cells and the
heading term is thus weighted by $T_k$. The field objective sums
the per-step losses and adds a collision penalty on the final posterior:
\begin{equation}
\mathcal{L}_{\mathrm{np}} = \sum^K_{k=1}\ell_k(P_k,T_k)
\;+\; \lambda_c\,\Omega(P_K),
\end{equation}
Each $k$-th refinement stage receives its own term as above, and at inference we read the final posterior $P_K$.

$P_t$ acts as the goal-conditioned estimate of a distribution whose mass encodes where paths to $g_t$ travel.
At test-time, the occupancy is also updated to raise the cost of frontier regions if a collision is detected, thus reducing repeated collisions.  

\subsection{Generator $f_{\mathrm{gen}}$: trajectory shaping}
Similar to \cite{wangSanDPlannerSampleEfficientDiffusion2026}, we use B-splines as the trajectory parameterization due to their $C^2$ continuity and favorable geometric properties. Unlike existing methods, we derive an endpoint from the egress predictor \textit{prior} to trajectory generation. Thus, we define the B-spline with $M$ interior control points as parameterized residually around the straight line to the egress point, $\tau = \tau_0(g_t) + B\delta$ where $\delta \in \mathbb{R}^{M \times 2}$. We use a standard DDPM \cite{hoDenoisingDiffusionProbabilistic2020} to denoise only the residuals around the straight line. The generator predicts the offset $\hat\delta_0 = f_{\mathrm{gen}}(\delta_i, i, c)$ where $c = \{P_t, g_t, v_t\}$ is the conditioning and $v_t$ is the velocity derived from the localization. We train the generator to denoise over 10 steps with the standard $x_0$-prediction plus a collision cost:
\begin{equation}
\mathcal{L}_{\mathrm{gen}}
= \mathbb{E}_{(\delta_0,\,c) \sim \mathcal{D}, \epsilon}\big[\,
\|\hat \delta_0-\delta_0\|^2  + \lambda_c \; \mathrm{coll}(C_t,\tau_0 + B \hat \delta_0)
\big] \,
\label{eq:gen}
\end{equation}
where $\mathrm{coll}(C,\tau) \;=\; \frac{1}{|\tau|}\sum_{x\in\tau} C(x)$.
The learning targets $\delta_0$ are drawn from the pre-computed $f_{\mathrm{np}}$ posterior predictions rather than the ground-truth; the generator learns to sample from the posterior probability. 
Given a frame's posterior field, we sample $N$ stochastic paths over posterior's high-probability corridor from origin to the subgoal, which are subsequently converted to B-spline control points. At generation time, we denoise a candidate set of samples $\mathcal{T}_N$ from Gaussian noise.

\begin{table*}[t]\centering\caption{Quantitative point-goal navigation results on 6060 episodes, across 60 diverse environments.}\label{tab:main}
{\setlength{\tabcolsep}{8pt}
\begin{tabular}{@{}l | l |cc|cc|cc|cc| c@{}}\toprule
 & Params. (train/total) & \multicolumn{2}{c}{Home} & \multicolumn{2}{c}{Commercial} & \multicolumn{2}{c}{Clut.-Easy} & \multicolumn{2}{c}{Clut.-Hard} & Coll.\\
 & & SR $\uparrow$ & SPL $\uparrow$ & SR $\uparrow$ & SPL $\uparrow$ & SR $\uparrow$ & SPL $\uparrow$ & SR $\uparrow$ & SPL $\uparrow$ & /10m $\downarrow$\\ \midrule
NavDP & 135.7M / 135.7M & \textbf{57.2} & \textbf{52.2} & \textbf{72.7} & \textbf{68.4} & \textbf{93.4} & \textbf{90.7} & \textbf{89.3} & \textbf{86.1} & {0.200}\\
ViPlanner & 71.0M / 115.1M  & 42.6 & 40.8 & 64.3 & 62.2 & 81.4 & 81.0 & 60.6 & 60.3 & 0.237\\
iPlanner & 53.3M / 53.3M & 39.5 & 37.5 & 53.9 & 52.1 & 90.3 & \underline{89.0} & 82.4 & \underline{80.8} & {0.173}\\ \midrule
Ours & \textbf{0.58M} / \textbf{22.7M} & \underline{51.4} & \underline{45.8} & \underline{70.6} & \underline{64.7} & \underline{91.6} & {84.4} & \underline{84.8} & {77.6} & \textbf{0.111}\\
\textcolor{black!60}{Ours w/ MBNetv2} &
\textcolor{black!60}{0.58M / 1.12M} &
\textcolor{black!60}{49.6} &
\textcolor{black!60}{44.6} &
\textcolor{black!60}{68.4} &
\textcolor{black!60}{62.9} &
\textcolor{black!60}{87.6} &
\textcolor{black!60}{79.7} &
\textcolor{black!60}{79.7} &
\textcolor{black!60}{70.7} &
\textcolor{black!60}{\underline{0.118}}\\
\bottomrule\end{tabular}}\end{table*}

\subsection{Critic}
As the diffusion model naturally generates diverse paths, some may be worse than others. We use the critic, a post-hoc deterministic selection over the candidate set minimizing
\begin{equation}
\tau^\star = \arg\min_{\tau \in \mathcal{T}_N} J(\tau), \qquad
J(\tau) = \frac{1}{|\tau|}\sum_{x \in \tau} e^{-d(x)/\lambda_d},
\label{eq:sel}
\end{equation}
where $J(\cdot)$ is essentially the costmap $C$ without the bearing ramp, to select the best candidate.

\section{Experiments}\label{sec:exp}
\subsection{Implementation Details}
\textbf{Training.} We sampled around 44k random single-frame states across 55 Matterport3D \cite{Matterport3D} scenes, where each state consists of an egocentric view over the ground-truth trajectory, and the short horizon (0-6m) trajectory. All the learnable modules cumulatively train in under an hour on a single RTX 4090, including one-time caching of features.

\textbf{Evaluation.} We evaluate our model and measure the performance of NavDP \cite{caiNavDPLearningSimtoReal2025}, ViPlanner \cite{roth2024viplanner} and iPlanner \cite{Yang-RSS-23} using their publicly released weights, on the comprehensive InternRobotics benchmarks \cite{caiNavDPLearningSimtoReal2025}, which consist of 4040 episodes in InternScenes and 2020 episodes in ClutteredEnv, spanning 60 different environments. We use the same PointNav evaluation protocol as the other works, where task performance is quantified by Success Rate (SR) and Success weighted by Path Length (SPL) \cite{andersonEvaluationEmbodiedNavigation2018}, which indicate the episodic binary goal-reaching state and divided by the path length over the ground-truth respectively.

\subsection{Main Results}
\begin{figure}[t]
  \centering
  \setlength{\lineskip}{-5pt}
  \begin{subfigure}{\columnwidth}
    \includegraphics[width=\columnwidth]{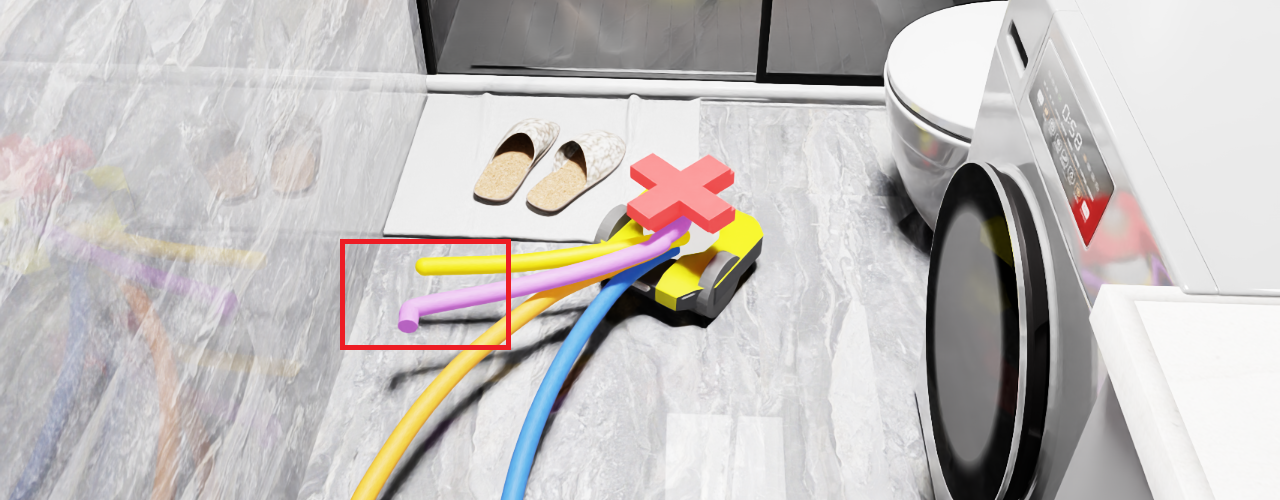}%
    \llap{\raisebox{2pt}{\colorbox{white}{\footnotesize(a)}}\hspace{2pt}}
    \label{fig:qual-bathroom}
  \end{subfigure}\\[-5pt]
  \begin{subfigure}{\columnwidth}
    \includegraphics[width=\columnwidth]{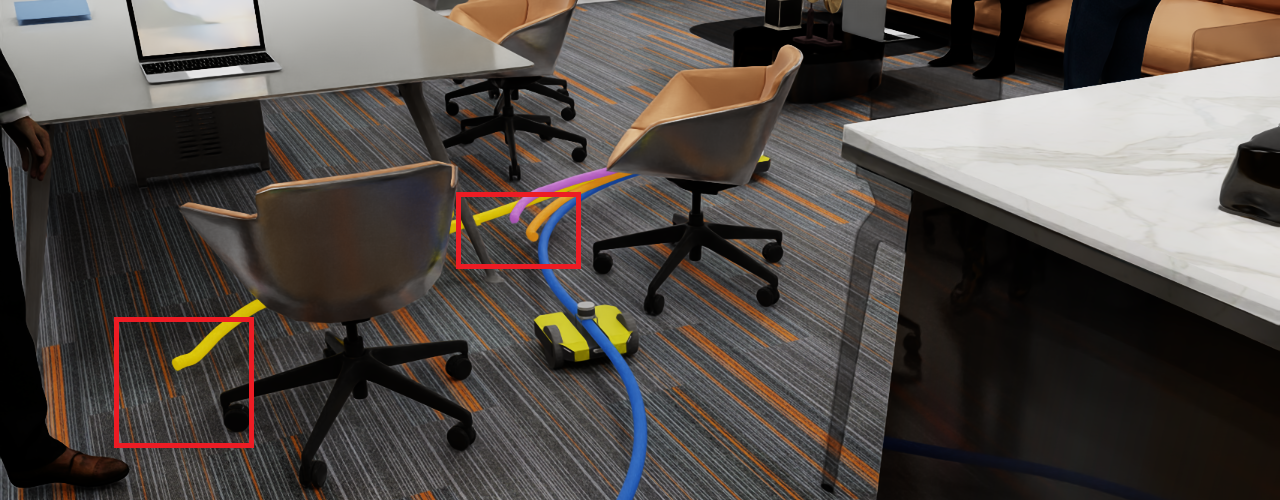}%
    \llap{\raisebox{2pt}{\colorbox{white}{\footnotesize(b)}}\hspace{2pt}}
    \label{fig:qual-cluttered}
  \end{subfigure}\\[-5pt]
  \begin{subfigure}{\columnwidth}
    \includegraphics[width=\columnwidth]{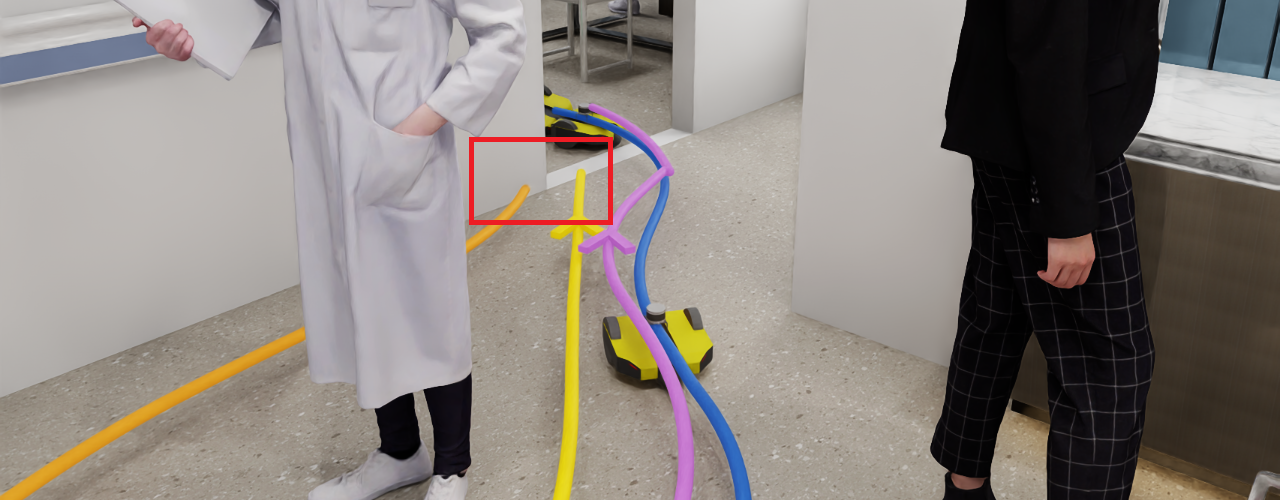}%
    \llap{\raisebox{2pt}{\colorbox{white}{\footnotesize(c)}}\hspace{2pt}}
    \label{fig:qual-hospital}
  \end{subfigure}
  \caption{\textbf{Qualitative comparison in simulation} showing \hl{ourblue}{ours} successfully navigating difficult conditions. Areas indicated in \textcolor{red}{red} denote episode failure points. In (a), suboptimal initialization position causes \hl{viyellow}{ViPlanner} and \hl{imagenta}{iPlanner} to collide with the wall, in (b) all others fail to navigate the tight space, and in (c) \hl{navorange}{NavDP} takes a suboptimal route, resulting in too sharp of an approach.}
  \label{fig:qual}
\end{figure}

Our system attains competitive results (see Table~\ref{tab:main}), outperforming two recent policies (ViPlanner, iPlanner) on average SR. In the realistic home and commercial scenes, our method trails NavDP by 2-6\% SR while having 233x fewer trainable parameters. Furthermore, our collision rate (indicated as collisions per 10m) is significantly lower than all other evaluated works, demonstrating that our system does not trade-off parameters for lower safety. We provide some qualitative renders of successful cases in Figure~\ref{fig:qual}. We also indicate our results when using a significantly smaller MobileNetv2 image encoder, and provide a detailed ablation in Section~\ref{subsec:generalize}.

\textbf{Collision safety. } We evaluate detailed collision results in Table~\ref{tab:coll}.
Over all 6060 episodes, our method has the lowest average collision rate  (Coll. / 10m), hard collisions (forces exceeding 25N), and the fewest episodes with collisions.

We also analyze the failure characteristics: ours has a substantially lower collision rate; it tends to navigate for a longer distance before colliding with an obstacle. Furthermore, examining the distribution of collisions across failed versus successful episodes reveals that all methods perform similarly, with the majority of collisions occurring within failed episodes.

\begin{table}[h]\centering\caption{Collision statistics on all episodes, 15N impact threshold.}\label{tab:coll}
\resizebox{\columnwidth}{!}{%
\begin{tabular}{@{}l|c|c|c|c@{}}\toprule
 & \textbf{Ours} & \textbf{ViPlanner} & \textbf{iPlanner} & \textbf{NavDP}\\ \midrule
Avg. SR $\uparrow$ & \underline{70.1} & 59.3 & 59.9 & \textbf{73.8}\\
Avg. Path Length (m) & 5.34 & 3.93 & 3.87 & 5.27 \\
\midrule
Coll. / 10\,m $\downarrow$ & \textbf{0.111} & 0.237 & \underline{0.173} & 0.200\\
\makebox[2em][c]{$\llcorner$}Hard ($>$25N) $\downarrow$ & \textbf{0.070} & 0.155 & \underline{0.111} & 0.118\\
Impact-episodes $\downarrow$ & \textbf{445} & 523 & \underline{463} & 579\\
Fail vs succ. impact* & 17.4/3.1\% & 15.7/3.8\% & 13.8/3.5\% & 24.4/4.3\%\\\bottomrule\end{tabular}
}
\\[2pt]\footnotesize *denotes the percentage of episodes with collisions, subdivided by whether the robot successfully navigated to the goal.
\end{table}

\textbf{Sensor robustness. } In our system, the input depth and RGB enter separately; the egress predictor consumes the RGB while the analytic BEV projection operation consumes the input depth. The diffusion model never sees either input directly. Thus, we investigate our method's robustness to simulated sensor noise and corruption. 

\begin{figure}[t]
    \centering
    \begin{subfigure}[b]{0.48\columnwidth}
        \centering
        \includegraphics[width=\textwidth]{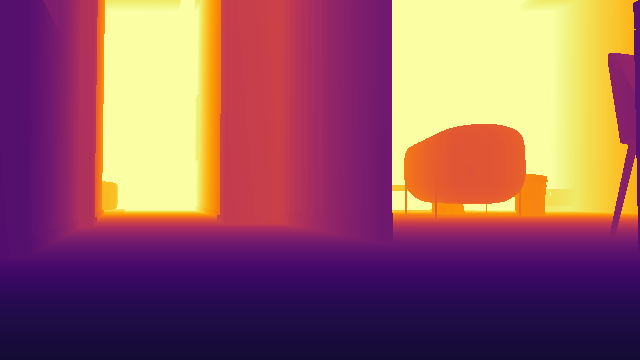} 
        \caption{Clean depth}
        \label{fig:row1_left}
    \end{subfigure}
    \hfill 
    \begin{subfigure}[b]{0.48\columnwidth}
        \centering
        \includegraphics[width=\textwidth]{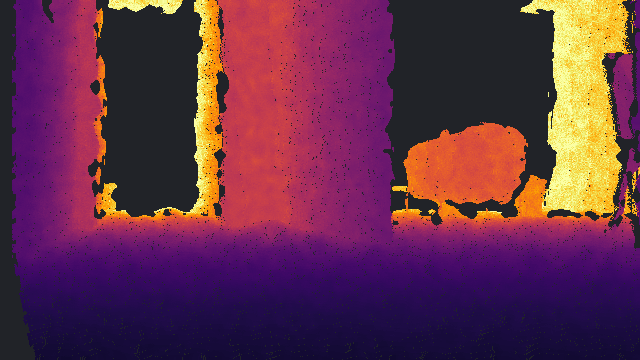}
        \caption{Simulated depth sensor noise}
        \label{fig:row1_right}
    \end{subfigure}
    \vspace{0pt} 
    \begin{subfigure}[b]{0.48\columnwidth}
        \centering
        \includegraphics[width=\textwidth]{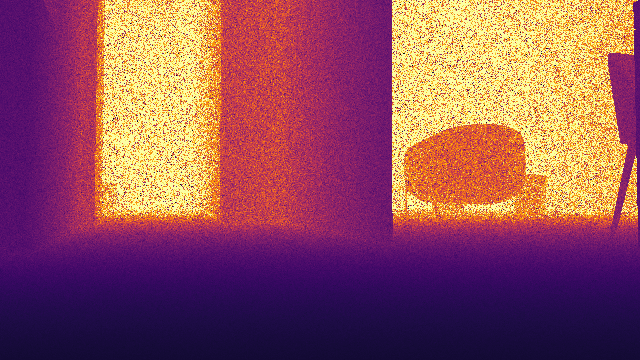}
        \caption{Depth with Gaussian noise}
        \label{fig:row2_left}
    \end{subfigure}
    \hfill
    \begin{subfigure}[b]{0.48\columnwidth}
        \centering
        \includegraphics[width=\textwidth]{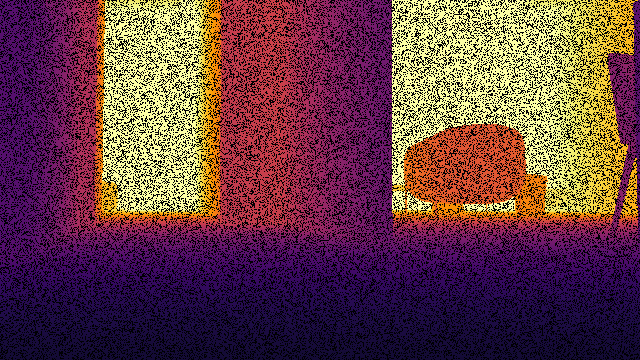}
        \caption{Depth with dropout}
        \label{fig:row2_right}
    \end{subfigure}
    \caption{Illustration of various simulated sensor corruption modes that were tested.}
    \label{fig:depth_grid}
\end{figure}

We evaluate robustness to depth-sensor degradation using a simplified noise model, based on the characteristics \cite{ahnAnalysisNoiseModeling2019} of the Intel RealSense D435, shown in Figure~\ref{fig:depth_grid}(b). We evaluate the performance degradation of our system on the same test suite in Table~\ref{tab:main}, and report the results in Table~\ref{tab:robust_depth}, where we show that our model is largely unaffected, while the other methods tend to show general performance degradation. All methods are tested zero-shot without any depth pre-processing or filters. We attribute marginal sub-\% SR increases in certain cases to a combination of enlarged obstacle footprint (i.e., from depth to occupancy projection), and run-to-run variance.

\begin{table}[h]\centering\caption{Simulated depth sensor corruption robustness, denoted as SR (and relative $\Delta$\%SR) $\uparrow$ compared to clean inputs.}\label{tab:robust_depth}
\resizebox{\columnwidth}{!}{%
\begin{tabular}{@{}l|c|c|c|c@{}}\toprule \textbf{} & \textbf{Home} & \textbf{Commercial} & \textbf{Clut.-Easy} & \textbf{Clut.-Hard}\\ \midrule
\textbf{Ours} & 51.4 (-0.0\%) & 71.6 ({+1.5\%}) & 93.2 ({+1.7\%}) & 86.3 ({+1.9\%})\\
iPlanner & 39.9 ({+0.9\%}) & 53.7 ({-0.2\%}) & 79.6 (\textcolor{red}{-11.8\%}) & 67.4 (\textcolor{red}{-18.1\%}) \\
ViPlanner & 36.9 (\textcolor{red}{-13.5\%}) & 56.5 (\textcolor{red}{-7.8\%}) & 25.6 (\textcolor{red}{-68.5\%}) & 17.1 (\textcolor{red}{-71.7\%}) \\
NavDP & 54.4 (\textcolor{red}{-4.9\%}) & 67.1 (\textcolor{red}{-7.7\%}) & 93.3 (-0.1\%) & 88.6 (\textcolor{red}{-1.1\%}) \\
\bottomrule
\end{tabular}%
}
\end{table}

We further test our method on various types of synthetic noise. Examples of the sensor corruption are shown in Figure~\ref{fig:depth_grid}. In Table~\ref{tab:robust}, we observe that our model is highly robust to RGB corruption (as a result of the pre-trained image encoder), and generally more robust than NavDP to depth corruption. However, ours fails under significant injected Gaussian noise of $\sigma=0.05z^2$, which we attribute to a failure of the analytic BEV calculation. As with classical methods, the failure mode is transparent and cheaply fixable; we recover baseline performance with a 3x3 median filter applied to the input depth. We evaluated performance on a subset of InternScenes-Home (101 episodes).

\begin{table}[t]\centering\caption{Sensor corruption robustness, denoted as relative SR (\%) $\uparrow$ compared to clean inputs.}\label{tab:robust}
\begin{tabular}{@{}l|c|c@{}}\toprule \textbf{Corruption} & \textbf{Ours} & \textbf{NavDP}\\ \midrule
Depth noise $\sigma{=}0.02z^2$ & \textbf{100} & 95\\
Depth noise $\sigma{=}0.05z^2$ & 19 & \textbf{92}\\
\makebox[2em][c]{$\llcorner$}3$\times$3 median filter & \textbf{100} & 92\\
Depth dropout 10\% & \textbf{100} & 95\\
Depth dropout 30\% & \textbf{91} & 61\\
RGB dark $\times$0.4 & \textbf{100} & 95\\
RGB noise $\sigma{=}20$ & \textbf{100} & 95\\ \bottomrule\end{tabular}\end{table}



\subsection{Interface Generalization}
\label{subsec:generalize}
\textbf{Other Goal Types. } In order to test how well our defined interfaces generalize, we evaluate our work on the No-Goal task in Table~\ref{tab:nogoal}, which measures how well a robot explores an unmapped area without collisions. In this experiment, we retrain only the egress head $f_\mathrm{eg}$ (around 123k parameters) by zeroing the goal conditioning on the same MP3D data, and reuse the other components $f_\mathrm{np}$, $f_\mathrm{gen}$ from the point-goal task without modification. For fairness, we also remove our generator's velocity conditioning and obstacle collision indicators in the costmap, since undirected navigation does not presume localization. We compare to NavDP's native no-goal exploration, observing that ours performs similarly on exploration area, tested on 100 respawn episodes in InternScenes-Home, with the termination condition set to collision (15N). This shows that our subgoal interface accurately preserves the necessary information for effective undirected navigation.

\begin{table}[h]\centering\caption{No-goal exploration in a randomly chosen scene, averaged over 100 trials.}\label{tab:nogoal}
\begin{tabular}{@{}l | cc | c@{}}\toprule
 & \multicolumn{2}{c}{\textbf{Explored area (m$^2$)}} & \textbf{Time}\\
 & \textbf{mean} $\uparrow$ & \textbf{median} $\uparrow$& \textbf{(s)} \\ \midrule
NavDP & 18.2 & 18.5 & 17.2\\
Ours & \textbf{18.3} & \textbf{18.8} & 25.7\\
\bottomrule\end{tabular}\end{table}

\textbf{Encoder \& Parameter Scaling. } To understand the significance of the image encoder, we truncate the pre-trained encoder at multiple points and retrain the egress operator (see Section~\ref{subsec:egress}) on the features at the truncated output. Since the interface with the rest of the model is the singular subgoal pixel, we reuse the rest of the model without training. We report the results on the hardest suite, InternScenes-Home (over 2k episodes) in Table~\ref{tab:enctrunc}.

We find that our proposed decomposition has minimal performance reduction even with significantly fewer parameters in the encoder, losing only 2.0 SR when using the much smaller MobileNetv2 \cite{sandlerMobileNetV2InvertedResiduals2019} (truncated at layer 13) as encoder. This demonstrates that the decomposition retains most of its navigation performance with a small ImageNet-pretrained backbone, requiring neither depth-specific priors nor large-scale self-supervised features.

\begin{table}[h]\centering\caption{Encoder ablation on InternScenes-Home. $\Delta$SR is relative to the full 12-block DAv2 ViT-S.}\label{tab:enctrunc}
\begin{tabular}{@{}l | c | c | cc @{}}\toprule
\textbf{Model} & \textbf{Blocks} & \textbf{Params. (M)} & \textbf{SR} $\uparrow$ & $\Delta$\textbf{SR} \\ \midrule
\multirow{5}{*}{DepthAnythingv2} & All (Baseline) & 22.06 & \textbf{51.4} & -- \\
 & 9  & 18.51 & 51.2 & -0.2 \\
 & 4  & 7.85  & 51.3 & -0.1 \\
 & 3  & 6.08  & 50.1 & -1.3 \\
 & 2  & 4.30  & 48.8 & -2.6 \\
 \midrule
 DINOv2 & All & 22.06 & 51.3 & -0.1\\
 \midrule
 MobileNetv2 & All & 0.54 & 49.6 & -1.8 \\
\bottomrule\end{tabular}\end{table}

\begin{figure*}[t]
  \centering
  \includegraphics[width=0.96\textwidth]{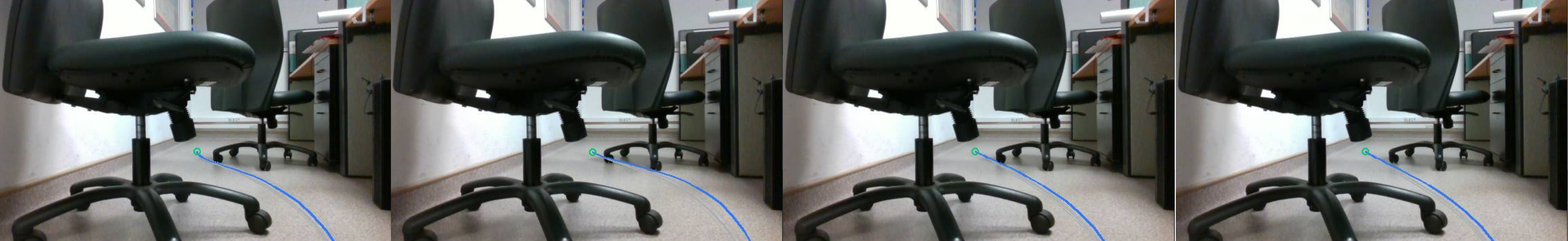}
  \caption{\textbf{Representative comparison of trajectories generated by different image encoders.} We visualize the generated trajectories through the same real-world scene, changing only the encoder as indicated in Table~\ref{tab:enctrunc}. From left to right: DepthAnythingv2 (3 blocks), DepthAnythingv2 (6 blocks), MobileNetv2, DepthAnythingv2 (Full).}
  \label{fig:abl-traj}
\end{figure*}

\subsection{Efficiency} Our system's end-to-end latency is approx. 9.17\,ms (109\,Hz) on a single RTX 4090, the majority of which is the image encoder. During execution in Isaac Sim, inference is capped in the planning thread at 10Hz (typ. $\sim$7Hz with overheads), so that our higher replan frequency does not unfairly bias the result in our favor. For inference, our model runs at $\sim$20Hz on Jetson Orin Nano without optimization (i.e., FP32 inference); our model runs on-device at the same or greater rate as the experiments in Table~\ref{tab:main}.

\begin{figure}[h]
  \centering
  \includegraphics[width=\columnwidth]{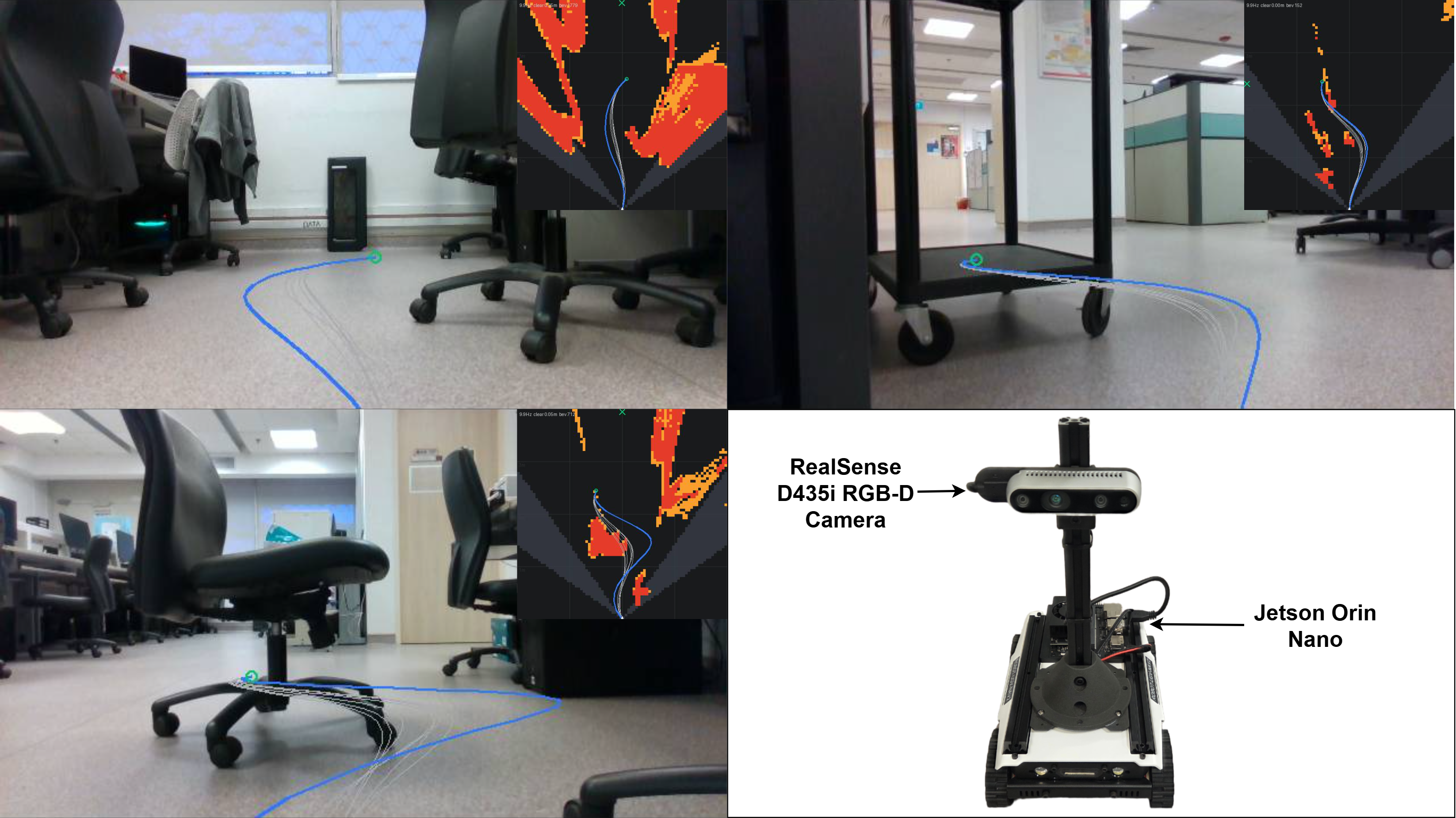}
  \caption{\textbf{Real-world deployment.} We deploy our system on a low-cost UGV (bottom-right quadrant). Our method generates feasible trajectories under real-world depth sensor characteristics. The set of generated trajectories is overlayed on the egocentric views, in grey, while the selected trajectory is in blue. The BEV-occupancy window is shown in the top-right of each quadrant, that also indicates the point-goal location as an 'X'.}
  \label{fig:realworld}
\end{figure}

\subsection{Real-world Experiments}
We test our system's zero-shot sim-to-real capability on a low-cost UGV (pictured in Figure~\ref{fig:realworld}), with a Jetson Orin Nano and Intel RealSense D435i RGB-D camera. All compute runs on the Jetson in real-time, including both our navigation model and cuVSLAM (RGB-D) for localization, at 10Hz and 30Hz respectively, both rates capped in software. We test the system in typical indoor environments (e.g., office, corridor) with presence of some dynamic objects. We test the system using challenging obstacles such as the office chairs pictured in Figure~\ref{fig:realworld}, which can be challenging for depth sensors. We also test in outdoor environments, pictured in Figure~\ref{fig:hero}   (left), such as long corridors and shared public spaces, with varying illumination conditions. These tests show that the on-device inference and navigation capabilities transfer to real-world environments, even with sub-optimal depth sensing and non-oracle localization. We provide qualitative trials, including those visualized in this paper, as Supplementary Materials.

\subsection{Ablation Studies}
In this section, we demonstrate that each of the key components of our method are load-bearing. In Table~\ref{tab:abl}, we tested two configurations: removing the egress predictor, instead feeding the downstream modules with the input goal, and removing the navigation predictor, training the generator on the depth-derived occupancy map instead of the posterior map. In both cases, SR is significantly lower. 

\begin{table}[h]\centering\caption{Ablation study on system components (on 101 episodes).}\label{tab:abl}
\begin{tabular}{@{}l|c|c@{}}\toprule \textbf{Configuration} & \textbf{SR} $\uparrow$ & \textbf{SPL} $\uparrow$\\ \midrule
Full system& \textbf{58.4} & \textbf{52.0}\\
\midrule
No navigation predictor & 3.0 & 3.0\\
No egress predictor & 20.8 & 20.4\\
\bottomrule\end{tabular}\end{table}

Next, we ablate several of our design decisions in Table~\ref{tab:abl2}. The first is \emph{sampling} (with replacement) from the egress predictor, instead of taking the soft $\arg\max$. This performs marginally worse: and our findings are that in most cases, the egress prediction is uni-modal (i.e., the mass concentrates around one pixel). In specific difficult scenes where the egress prediction has a multimodal distribution, the sampling improves SR by acting as a tie-breaker, but on average reduces performance.

The second is training the generator against ground-truth trajectories, instead of from randomly drawn samples from the posterior. We find that in this configuration, the model performs poorly, which we once again attribute to preservation of multimodality, where in density space (i.e., the posterior), mode-averaging is benign, while in the trajectory space, mode-averaging can lead to catastrophically wrong predictions.

The third is to train the modules end-to-end, which we demonstrate significantly degrades SR. For the $\arg\max$, we used soft $\arg\max$ in the implementation, and also tried discretization with Gumbel STE; both had similar results.

We then ablate various supporting components: the velocity conditioning in the DDPM, and the collision detection, which, when removed, reduce SR slightly, but are otherwise not significant.

Lastly, we evaluate an A* classical planner on the costmap $C_t$, which performs poorly. This demonstrates that the neural modules (and specifically the generator) perform non-trivial work over a classical planner.

\begin{table}[t]\centering\caption{Ablation study on design decisions (on 101 episodes).}\label{tab:abl2}
\begin{tabular}{@{}l|c|c@{}}\toprule \textbf{Configuration} & \textbf{SR} $\uparrow$ & \textbf{SPL} $\uparrow$\\ \midrule
Full system & \textbf{58.4} & \textbf{52.0}\\
\midrule
Sampling instead of argmax & 54.5 & 50.2\\ 
Training DDPM on GT objectives & 34.7 & 33.9 \\ 
Joint end-to-end training & 35.6 & 34.6\\ 
\midrule
No velocity conditioning & 55.5 & 49.3\\ 
No collision detection & 54.6 & 50.8\\
\midrule
Classical planner on costmap & 21.8 & 21.6\\ 
\bottomrule\end{tabular}\end{table}

\section{Limitations and Future Work}
Although our method is relatively robust to depth noise, it is currently limited to navigating on one floor, due to the use of depth-derived BEV maps. To navigate across multiple floors, this representation would need to be adapted to use voxel maps or other 3D representations capable of modeling elevation discontinuities.
Future work may consider adapting our work to image-goal navigation, which remains untested, yet can potentially make use of our proposed interface.

\section{Conclusions}

In this work, we have demonstrated that with few parameters, a structured system can have competitive point-goal navigation performance compared to much larger models. We demonstrate this claim with an architecture that uses three learnable modules, each performing a sub-task contributing to trajectory generation: a local subgoal prediction, a goal-conditioned navigability field, and a spline-parameterized trajectory shape generator. We do this while maintaining low collision rates, robustness to simulated sensor noise, and demonstrate transferability to no-goal navigation tasks. Finally, we qualitatively demonstrated the method transfers zero-shot to real environments.

\bibliographystyle{IEEEtran} 
\bibliography{references}    

@misc{caiNavDPLearningSimtoReal2025,
    title = {{NavDP}: {Learning} {Sim}-to-{Real} {Navigation} {Diffusion} {Policy} with {Privileged} {Information} {Guidance}},
    copyright = {arXiv.org perpetual, non-exclusive license},
    shorttitle = {{NavDP}},
    url = {https://arxiv.org/abs/2505.08712},
    doi = {10.48550/ARXIV.2505.08712},
    urldate = {2026-05-16},
    publisher = {arXiv},
    author = {Cai, Wenzhe and Peng, Jiaqi and Yang, Yuqiang and Zhang, Yujian and Wei, Meng and Wang, Hanqing and Chen, Yilun and Wang, Tai and Pang, Jiangmiao},
    year = {2025},
    note = {Version Number: 3},
}

@misc{wangSanDPlannerSampleEfficientDiffusion2026,
    title = {{SanD}-{Planner}: {Sample}-{Efficient} {Diffusion} {Planner} in {B}-{Spline} {Space} for {Robust} {Local} {Navigation}},
    copyright = {Creative Commons Attribution 4.0 International},
    shorttitle = {{SanD}-{Planner}},
    doi = {10.48550/ARXIV.2602.00923},
    url = {https://arxiv.org/abs/2602.00923v1},
    urldate = {2026-07-07},
    publisher = {arXiv},
    author = {Wang, Jincheng and Bao, Lingfan and Yang, Tong and Plasencia, Diego Martinez and Jiao, Jianhao and Kanoulas, Dimitrios},
    year = {2026},
    note = {Version Number: 1},
}

@inproceedings{shah2022gnm,
    title = {{GNM}: a general navigation model to drive any robot},
    url = {https://arxiv.org/abs/2210.03370},
    booktitle = {International conference on robotics and automation ({ICRA})},
    author = {Shah, Dhruv and Sridhar, Ajay and Bhorkar, Arjun and Hirose, Noriaki and Levine, Sergey},
    year = {2023},
}

@inproceedings{shah2023vint,
    title = {{ViNT}: a foundation model for visual navigation},
    url = {https://arxiv.org/abs/2306.14846},
    booktitle = {7th annual conference on robot learning},
    author = {Shah, Dhruv and Sridhar, Ajay and Dashora, Nitish and Stachowicz, Kyle and Black, Kevin and Hirose, Noriaki and Levine, Sergey},
    year = {2023},
}

@inproceedings{Wijmans2020DD-PPO:,
    title = {{DD}-{PPO}: {Learning} near-perfect {PointGoal} navigators from 2.5 billion frames},
    url = {https://openreview.net/forum?id=H1gX8C4YPr},
    booktitle = {International conference on learning representations},
    author = {Wijmans, Erik and Kadian, Abhishek and Morcos, Ari and Lee, Stefan and Essa, Irfan and Parikh, Devi and Savva, Manolis and Batra, Dhruv},
    year = {2020},
}

@inproceedings{chi2023diffusionpolicy,
    title = {Diffusion policy: {Visuomotor} policy learning via action diffusion},
    booktitle = {Proceedings of robotics: {Science} and systems ({RSS})},
    author = {Chi, Cheng and Feng, Siyuan and Du, Yilun and Xu, Zhenjia and Cousineau, Eric and Burchfiel, Benjamin and Song, Shuran},
    year = {2023},
}

@misc{weiGroundSlowMove2025,
    title = {Ground {Slow}, {Move} {Fast}: {A} {Dual}-{System} {Foundation} {Model} for {Generalizable} {Vision}-and-{Language} {Navigation}},
    copyright = {Creative Commons Attribution 4.0 International},
    shorttitle = {Ground {Slow}, {Move} {Fast}},
    url = {https://arxiv.org/abs/2512.08186},
    doi = {10.48550/ARXIV.2512.08186},
    urldate = {2026-05-16},
    publisher = {arXiv},
    author = {Wei, Meng and Wan, Chenyang and Peng, Jiaqi and Yu, Xiqian and Yang, Yuqiang and Feng, Delin and Cai, Wenzhe and Zhu, Chenming and Wang, Tai and Pang, Jiangmiao and Liu, Xihui},
    year = {2025},
    note = {Version Number: 1},
}

@article{sridhar2023nomad,
    title = {{NoMaD}: {Goal} masked diffusion policies for navigation and exploration},
    url = {https://arxiv.org/abs/2310.07896},
    journal = {arXiv pre-print},
    author = {Sridhar, Ajay and Shah, Dhruv and Glossop, Catherine and Levine, Sergey},
    year = {2023},
}

@misc{luo2026stepnavstructuredtrajectorypriors,
    title = {{StepNav}: {Structured} trajectory priors for efficient and multimodal visual navigation},
    url = {https://arxiv.org/abs/2602.02590},
    author = {Luo, Xubo and Wu, Aodi and Han, Haodong and Wan, Xue and Zhang, Wei and Shu, Leizheng and Wang, Ruisuo},
    year = {2026},
    note = {arXiv: 2602.02590 [cs.RO]},
}

@inproceedings{ren2025prior,
    title = {Prior does matter: {Visual} navigation via denoising diffusion bridge models},
    booktitle = {Proceedings of the computer vision and pattern recognition conference},
    author = {Ren, Hao and Zeng, Yiming and Bi, Zetong and Wan, Zhaoliang and Huang, Junlong and Cheng, Hui},
    year = {2025},
    pages = {12100--12110},
}

@misc{luanRectifiedSchrodingerBridge2026,
    title = {Rectified {Schrödinger} {Bridge} {Matching} for {Few}-{Step} {Visual} {Navigation}},
    url = {http://arxiv.org/abs/2604.05673},
    doi = {10.48550/arXiv.2604.05673},
    urldate = {2026-07-10},
    publisher = {arXiv},
    author = {Luan, Wuyang and Li, Junhui and Zhao, Weiguang and Zhang, Wenjian and Wu, Tieru and Ma, Rui},
    month = may,
    year = {2026},
    note = {arXiv:2604.05673 [cs.RO]},
}

@inproceedings{roth2024viplanner,
    title = {Viplanner: {Visual} semantic imperative learning for local navigation},
    booktitle = {2024 {IEEE} international conference on robotics and automation ({ICRA})},
    publisher = {IEEE},
    author = {Roth, Pascal and Nubert, Julian and Yang, Fan and Mittal, Mayank and Hutter, Marco},
    year = {2024},
    pages = {5243--5249},
}

@inproceedings{Yang-RSS-23,
    address = {Daegu, Republic of Korea},
    title = {{iPlanner}: {Imperative} path planning},
    doi = {10.15607/RSS.2023.XIX.064},
    booktitle = {Proceedings of robotics: {Science} and systems},
    author = {Yang, Fan and Wang, Chen and Cadena, Cesar and Hutter, Marco},
    month = jul,
    year = {2023},
}

@article{Matterport3D,
    title = {{Matterport3D}: {Learning} from {RGB}-{D} data in indoor environments},
    journal = {International Conference on 3D Vision (3DV)},
    author = {Chang, Angel and Dai, Angela and Funkhouser, Thomas and Halber, Maciej and Niessner, Matthias and Savva, Manolis and Song, Shuran and Zeng, Andy and Zhang, Yinda},
    year = {2017},
}

@misc{andersonEvaluationEmbodiedNavigation2018,
    title = {On {Evaluation} of {Embodied} {Navigation} {Agents}},
    url = {http://arxiv.org/abs/1807.06757},
    doi = {10.48550/arXiv.1807.06757},
    urldate = {2026-07-11},
    publisher = {arXiv},
    author = {Anderson, Peter and Chang, Angel and Chaplot, Devendra Singh and Dosovitskiy, Alexey and Gupta, Saurabh and Koltun, Vladlen and Kosecka, Jana and Malik, Jitendra and Mottaghi, Roozbeh and Savva, Manolis and Zamir, Amir R.},
    month = jul,
    year = {2018},
    note = {arXiv:1807.06757 [cs.AI]},
}

@misc{hoDenoisingDiffusionProbabilistic2020,
    title = {Denoising {Diffusion} {Probabilistic} {Models}},
    url = {http://arxiv.org/abs/2006.11239},
    doi = {10.48550/arXiv.2006.11239},
    urldate = {2026-07-11},
    publisher = {arXiv},
    author = {Ho, Jonathan and Jain, Ajay and Abbeel, Pieter},
    month = dec,
    year = {2020},
    note = {arXiv:2006.11239 [cs.LG]},
}

@article{foxDynamicWindowApproach1997,
    title = {The dynamic window approach to collision avoidance},
    volume = {4},
    copyright = {https://ieeexplore.ieee.org/Xplorehelp/downloads/license-information/IEEE.html},
    issn = {10709932},
    url = {http://ieeexplore.ieee.org/document/580977/},
    doi = {10.1109/100.580977},
    number = {1},
    urldate = {2026-07-10},
    journal = {IEEE Robotics \& Automation Magazine},
    author = {Fox, D. and Burgard, W. and Thrun, S.},
    month = mar,
    year = {1997},
    pages = {23--33},
}

@inproceedings{lee2023adaptive,
    title = {Adaptive and explainable deployment of navigation skills via hierarchical deep reinforcement learning},
    booktitle = {2023 {IEEE} international conference on robotics and automation ({ICRA})},
    author = {Lee, Kyowoon and Kim, Seongun and Choi, Jaesik},
    year = {2023},
    pages = {1673--1679},
}

@inproceedings{NEURIPS2024_26cfdcd8,
    title = {Depth anything {V2}},
    volume = {37},
    doi = {10.52202/079017-0688},
    booktitle = {Advances in neural information processing systems},
    publisher = {Curran Associates, Inc.},
    author = {Yang, Lihe and Kang, Bingyi and Huang, Zilong and Zhao, Zhen and Xu, Xiaogang and Feng, Jiashi and Zhao, Hengshuang},
    editor = {Globerson, A. and Mackey, L. and Belgrave, D. and Fan, A. and Paquet, U. and Tomczak, J. and Zhang, C.},
    year = {2024},
    pages = {21875--21911},
}

@inproceedings{NEURIPS2020_2c75cf26,
    title = {Object goal navigation using goal-oriented semantic exploration},
    volume = {33},
    booktitle = {Advances in neural information processing systems},
    publisher = {Curran Associates, Inc.},
    author = {Chaplot, Devendra Singh and Gandhi, Dhiraj Prakashchand and Gupta, Abhinav and Salakhutdinov, Russ R},
    editor = {Larochelle, H. and Ranzato, M. and Hadsell, R. and Balcan, M.F. and Lin, H.},
    year = {2020},
    pages = {4247--4258},
}

@misc{pengLoGoPlannerLocalizationGrounded2025,
    title = {{LoGoPlanner}: {Localization} {Grounded} {Navigation} {Policy} with {Metric}-aware {Visual} {Geometry}},
    copyright = {arXiv.org perpetual, non-exclusive license},
    shorttitle = {{LoGoPlanner}},
    url = {https://arxiv.org/abs/2512.19629},
    doi = {10.48550/ARXIV.2512.19629},
    urldate = {2026-07-07},
    publisher = {arXiv},
    author = {Peng, Jiaqi and Cai, Wenzhe and Yang, Yuqiang and Wang, Tai and Shen, Yuan and Pang, Jiangmiao},
    year = {2025},
    note = {Version Number: 2},
}

@inproceedings{ahnAnalysisNoiseModeling2019,
    address = {Jeju, Korea (South)},
    title = {Analysis and {Noise} {Modeling} of the {Intel} {RealSense} {D435} for {Mobile} {Robots}},
    copyright = {https://ieeexplore.ieee.org/Xplorehelp/downloads/license-information/IEEE.html},
    isbn = {978-1-7281-3232-7},
    url = {https://ieeexplore.ieee.org/document/8768489/},
    doi = {10.1109/URAI.2019.8768489},
    urldate = {2026-08-31},
    booktitle = {2019 16th {International} {Conference} on {Ubiquitous} {Robots} ({UR})},
    publisher = {IEEE},
    author = {Ahn, Min Sung and Chae, Hosik and Noh, Donghun and Nam, Hyunwoo and Hong, Dennis},
    month = jun,
    year = {2019},
    pages = {707--711},
}

@misc{sandlerMobileNetV2InvertedResiduals2019,
    title = {{MobileNetV2}: {Inverted} {Residuals} and {Linear} {Bottlenecks}},
    shorttitle = {{MobileNetV2}},
    url = {http://arxiv.org/abs/1801.04381},
    doi = {10.48550/arXiv.1801.04381},
    urldate = {2026-09-05},
    publisher = {arXiv},
    author = {Sandler, Mark and Howard, Andrew and Zhu, Menglong and Zhmoginov, Andrey and Chen, Liang-Chieh},
    month = mar,
    year = {2019},
    note = {arXiv:1801.04381 [cs.CV]},
}

\newpage

\vfill

\end{document}